\pdfoutput=1

\documentclass[11pt]{article}

\usepackage{ACL2023}

\usepackage{times}
\usepackage{latexsym}

\usepackage[T1]{fontenc}

\usepackage[utf8]{inputenc}

\usepackage{microtype}

\usepackage{inconsolata}
\usepackage{graphicx}
\usepackage{booktabs}
\usepackage{amsmath}
\usepackage{arydshln}
\usepackage{subfigure}
\usepackage{CJKutf8}
\usepackage{multirow}
\usepackage{float}

\title{Rethinking Masked Language Modeling for Chinese Spelling Correction}

\author{Hongqiu Wu\textsuperscript{\rm 1,2,\thanks{\;\,Work was done during a cooperation with ByteDance.}} \and Shaohua Zhang\textsuperscript{\rm 3} \and Yuchen Zhang\textsuperscript{\rm 3} \and Hai Zhao\textsuperscript{\rm 1,2,\thanks{\;\,Corresponding author; This paper was partially supported by Key Projects of National Natural Science Foundation of China (U1836222 and 61733011).}} \\
        \textsuperscript{\rm 1}Department of Computer Science and Engineering, Shanghai Jiao Tong University \\
        \textsuperscript{\rm 2}Key Laboratory of Shanghai Education Commission for Intelligent Interaction \\
        and Cognitive Engineering, Shanghai Jiao Tong University \\
        \textsuperscript{\rm 3}ByteDance \\
        \texttt{wuhongqiu@sjtu.edu.cn,zhang.shaohua.cs@gmail.com} \\
        \texttt{zhangyuc@gmail.com,zhaohai@cs.sjtu.edu.cn}}

\begin{document}
\maketitle
\begin{abstract}

In this paper, we study Chinese Spelling Correction (CSC) as a joint decision made by two separate models: a language model and an error model. Through empirical analysis, we find that fine-tuning BERT tends to over-fit the error model while under-fit the language model, resulting in poor generalization to out-of-distribution error patterns. Given that BERT is the backbone of most CSC models, this phenomenon has a significant negative impact. To address this issue, we are releasing a multi-domain benchmark \textit{LEMON}, with higher quality and diversity than existing benchmarks, to allow a comprehensive assessment of the open domain generalization of CSC models. Then, we demonstrate that a very simple strategy – randomly masking 20\% non-error tokens from the input sequence during fine-tuning – is sufficient for learning a much better language model without sacrificing the error model. This technique can be applied to any model architecture and achieves new state-of-the-art results on SIGHAN, ECSpell, and LEMON\footnote{\url{https://github.com/gingasan/lemon}}.

\end{abstract}

\section{Introduction}

\begin{figure}
\centering
\includegraphics[width=0.46\textwidth]{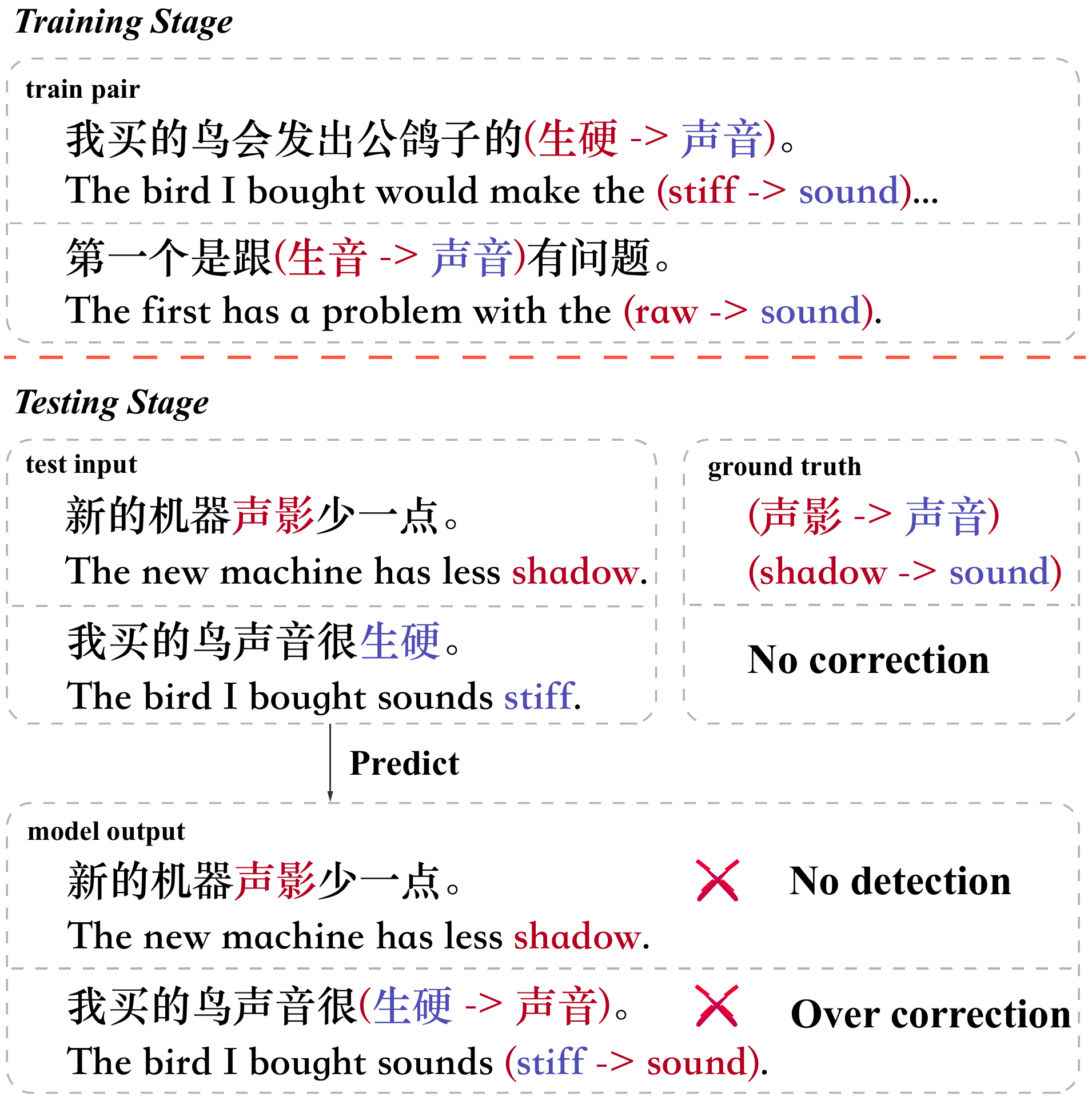}
\caption{Mistakes made by regularly fine-tuned BERT.}
\label{ex}
\end{figure}

Chinese Spelling Correction (CSC) is a crucial task in natural language processing (NLP) behind many downstream applications, e.g, web search \citep{DBLP:conf/tal/MartinsS04,DBLP:conf/coling/GaoLMQS10}, named entity recognition, optical character recognition \citep{DBLP:conf/lrec/AfliQWS16,DBLP:conf/emnlp/GuptaCBHB21}. It aims to detect and correct the potential spelling errors in a sentence. BERT \citep{DBLP:conf/naacl/DevlinCLT19} and its enhanced variants have achieved state-of-the-art results in the current CSC community (name a few) \citep{DBLP:conf/acl/ZhangHLL20,DBLP:conf/acl/LiuYYZW20,DBLP:conf/acl/ZhuYZM22}.

From a high-level perspective, CSC requires a \emph{language model} and an \emph{error model} working collaboratively to make a decision \citep{DBLP:conf/coling/KernighanCG90}. Suppose that the input sentence contains $n$ characters $X = (x_1,...,x_n)$. The model predicts the corrected character at each position $Y = (y_1,...,y_n)$. At each position $i$, let $x_{-i}$ indicate the characters at all other positions, then by Bayes Rule \citep{DBLP:conf/coling/KernighanCG90}, we have:
\begin{equation}\label{eqn:bayes}
P(y_i|X)\propto \underbrace{P(y_i|x_{-i})}_{\text{language model}} \cdot \underbrace{P(x_i|y_i,x_{-i})}_{\text{error model}}
\end{equation}
where the language model decides the distribution of the character $y_i$ given the context, while the error model represents the distribution of the potential misspelled character $x_i$ given the context and its correct form (see Appendix \ref{appendix:a} for the derivation). According to the BERT architecture, these two models are jointly trained and evaluated. However, their respective performances have not been throughout studied by previous work.

\begin{CJK}{UTF8}{gkai}
In this paper, we make a key observation that BERT-based CSC models typically over-fit the error model, yet under-fit the language model, because the error model is much easier to memorize compared to the language model. As a result, the model generalizes very poor to unseen \textit{edit pairs} $(x_i,y_i)$ and fails to exploit the context $x_{-i}$. We illustrate this fact in Figure \ref{ex}. Here, the model has been exposed to edit pairs ``生硬$\rightarrow$声音'' (correct \textit{stiff} to \textit{sound}) and ``生音$\rightarrow$声音'' (correct \textit{raw} to \textit{sound}) during training. During testing, the model fails to detect an unseen edit pair ``声影$\rightarrow$声音'' (correct \textit{shadow} to \textit{sound}) and meanwhile over-corrects ``生硬$\rightarrow$声音'' (correct \textit{stiff} to \textit{sound}). This is due to the fact that the model naively memorizes the training edit pairs, failing to identify if they fit the broader context.
We will present qualitative analysis of this phenomenon in later sections.
\end{CJK}

The consequence of a sub-optimal or under-fit language model is that the model struggles to generalize to new contexts and new domains.
SIGHAN is the current most widely-used benchmark in CSC, but it is limited in two ways: (1) a narrow sentence corpus sourced exclusively from the Chinese essays by foreign speakers \citep{DBLP:conf/acl-sighan/WuLL13}; (2) a low diversity of edit pairs (i.e. 370 edit pairs in its test set). As a result, it does not pose enough challenge to the model's generalization ability. To this end, we present \textit{LEMON}, a new benchmark that is a \textit{\textbf{l}arge-scal\textbf{e} \textbf{m}ulti-d\textbf{o}main dataset with \textbf{n}}atural spelling errors, which spans 7 domains and contains over 22,000 examples with 7,627 distinct edit pairs collected from real human daily writing. It provides a comprehensive evaluation of CSC models in real-world scenarios.

Based on LEMON and other public benchmarks, we demonstrate that a very simple method can effectively enhance language modeling without causing adverse effect to error modeling, thus significantly improves CSC model performances. The method is to randomly mask 20\% of the non-error tokens from the input sentence during fine-tuning (this is different from masking 15\% tokens during pre-training in BERT). If $x_i$ is masked, it forces the model to predict $y_i$ given $x_{-i}$ without any clue about $x_i$, equivalent to training $P(y_i|x_{-i})$. This masked-fine-tuning (Masked-FT) technique is unlike other data augmentation methods based on homophone substitution, random substitution or confusion sets \citep{DBLP:conf/aaai/ZhaoW20,DBLP:conf/acl/LiuYYZW20}, in that it does not impose any assumption about human errors. As a result, it enables learning a completely unbiased error model from real human data. This property let Masked-FT achieve new state-of-the-art across CSC benchmarks.

We also show that Masked-FT is effective in domain transfer. Suppose that there is an annotated parallel corpus for a certain domain, and we want to transfer the model of such a domain to a new domain where only monolingual (i.e. unannotated) corpus is available. We propose to train the model with the parallel data along with a masked language modeling (MLM) loss from the monolingual corpus. The idea behind is to transfer the language model to the new domain while preserving the error model that is learned through the parallel data. Empirical results demonstrate that this way of using monolingual data produces a better model than data synthesis methods based on confusion sets.

Our contributions are summarized as follows. (1) We perform empirical analysis showing that BERT-based CSC models learn a sub-optimal language model, resulting in a bad performance on out-of-distribution edit pairs. (2) We release a large-scale and multi-domain benchmark for CSC, which is more challenging than existing ones. (3) We demonstrate that a simple masked-fine-tuning strategy significantly enhance language modeling without hurting error modeling, leading to new state-of-the-art results across benchmarks.

\section{Analysis of BERT fine-tuning}

In this section, we report empirical analysis on BERT-based models. We study their top-k performance, generalization to unseen edit pairs, and gradient scales during training. The observation is that the BERT-based models, with regular fine-tuning, easily over-fits the edit pairs in the training set and learns a degenerated language model. For some analyses, we also include the result of masked-FT (randomly mask 20\% input tokens) for comparative study.

\subsection{Top-k Predictions}

CSC typically cares about the top-1 prediction at each position. But here, we print out the top-5 predictions in order to get a sense of its language modeling capability. We find that the fine-tuned BERT model tends to predict homophones and homographs of the input character, regardless of its contextual appropriateness. Note that homophones and homographs are the two main forms of spelling errors in Chinese. Thus, it reveals that the error model has dominated the prediction. In contrast, the model trained with Masked-FT tends to predict characters that fits the context better.

\begin{CJK}{UTF8}{gkai}

We demonstrate two cases in Table \ref{topk}. In the first case, both models make the correct top-1 prediction. At top 2-5, however, the fine-tuned model predicts a list of homophones: ``年纪'', ``年机'' and ``年轻'', ``年青''. None of them makes any sense in the context. Masked-FT predicts ``年龄'', ``年岁'', and ``年代'', all carrying the meaning of \textit{age} in Chinese, which fits the context. In the second case, the fine-tuned model predicts the correct answer at top-4, but through top 2-3, the predictions ``景'' (a homograph of ``影'') and ``应'' (a homophone of ``影'') don't fit the context at all. In contrast, the Masked-FT model predicts ``声音'', ``声声'', and ``声响'', which all represent the correct meaning: \textit{sound}. All the homophones and homographs that the FT model predicts come from the popular edit pairs in the training data.

\begin{table}[]
\centering
\small
\setlength\tabcolsep{4pt}
\begin{tabular}{@{}ll@{}}
\toprule
\textit{source}& 吴阿姨年{\color[HTML]{DC143C}级}大了。 \\
\textit{FT}&吴阿姨年({\color[HTML]{00008B}纪},{\color[HTML]{00008B}级},{\color[HTML]{00008B}机},{\color[HTML]{00008B}轻},{\color[HTML]{00008B}青})大了。 \\
\textit{Masked-FT}&吴阿姨年({\color[HTML]{00008B}纪},{\color[HTML]{00008B}级},{\color[HTML]{00008B}龄},{\color[HTML]{00008B}岁},{\color[HTML]{00008B}代})大了。\\ \hline\hline
\textit{source}&新的机器有可能声{\color[HTML]{DC143C}影}少一点。 \\
\textit{FT}& 新的机器有可能声({\color[HTML]{00008B}影},{\color[HTML]{00008B}景},{\color[HTML]{00008B}应},{\color[HTML]{00008B}音},{\color[HTML]{00008B}引})少一点。 \\
\textit{Masked-FT}&新的机器有可能声({\color[HTML]{00008B}音},{\color[HTML]{00008B}影},{\color[HTML]{00008B}声},{\color[HTML]{00008B}响},{\color[HTML]{00008B}味})少一点。 \\ \bottomrule
\end{tabular}
\caption{Top-$k$ results each model recalls on the same sentence. The models here are trained on SIGHAN. \textit{FT} refers to regular fine-tuning.}
\label{topk}
\end{table}
\end{CJK}

\subsection{Seen vs. Unseen Edit Pairs}
\label{inc-exc}

\begin{table}[b]
\centering
\small
\setlength\tabcolsep{5pt}
\begin{tabular}{@{}l|l|lll@{}}
\toprule
                            &     & Prec.                      & Rec.                      & F1   \\ \hline\hline
\multirow{2}{*}{fine-tuned} & INC & 73.5                       & 56.8                      & 64.1 \\
                            & EXC & 10.7 $_{\downarrow 62.8}$  & 4.4 $_{\downarrow 52.4}$  & 6.3 $_{\downarrow 57.8}$ \\ \hline
\multirow{2}{*}{vanilla BERT }        & INC & 51.5                       & 48.5                      & 49.9 \\
                            & EXC & 46.3                       & 45.0                      & 45.6 \\ \bottomrule
\end{tabular}
\caption{CSC performance crash on unseen edit pairs.}
\label{prf}
\end{table}

In this experiment, we separate the test set of SIGHAN \citep{DBLP:conf/acl-sighan/TsengLCC15} into two subsets, INC (shorthand for \emph{inclusive}, representing edit pairs that overlap with the training set) and EXC (shorthand for \emph{exclusive}, with edit pairs that do not emerge in the training set). Table \ref{prf} shows the comparison. The fine-tuned BERT fits INC well (F1=64.1), but the performance sharply drops on EXC (F1=6.3). It suggests that the model generalizes poorly to unseen edit pairs where the error model does not provide any useful signal.

It is worth noting that for many unseen edit pairs, although they never appear in the training data, they can actually be corrected by human based on the context. To illustrate this fact, we attempt to utilize a vanilla BERT to correct the errors by replacing the misspelled token by \texttt{[MASK]}. Surprisingly, we find that the vanilla BERT can actually achieve a decent accuracy (F1=45.6) on EXC, much better than the fine-tuned BERT (F1=6.3). This result highlights the fact that a well-trained language model has a great potential to handle unseen error patterns.



\subsection{Gradient Norm}

\begin{figure}
\centering
\subfigure[Gradient Norm]{
\includegraphics[width=0.21\textwidth]{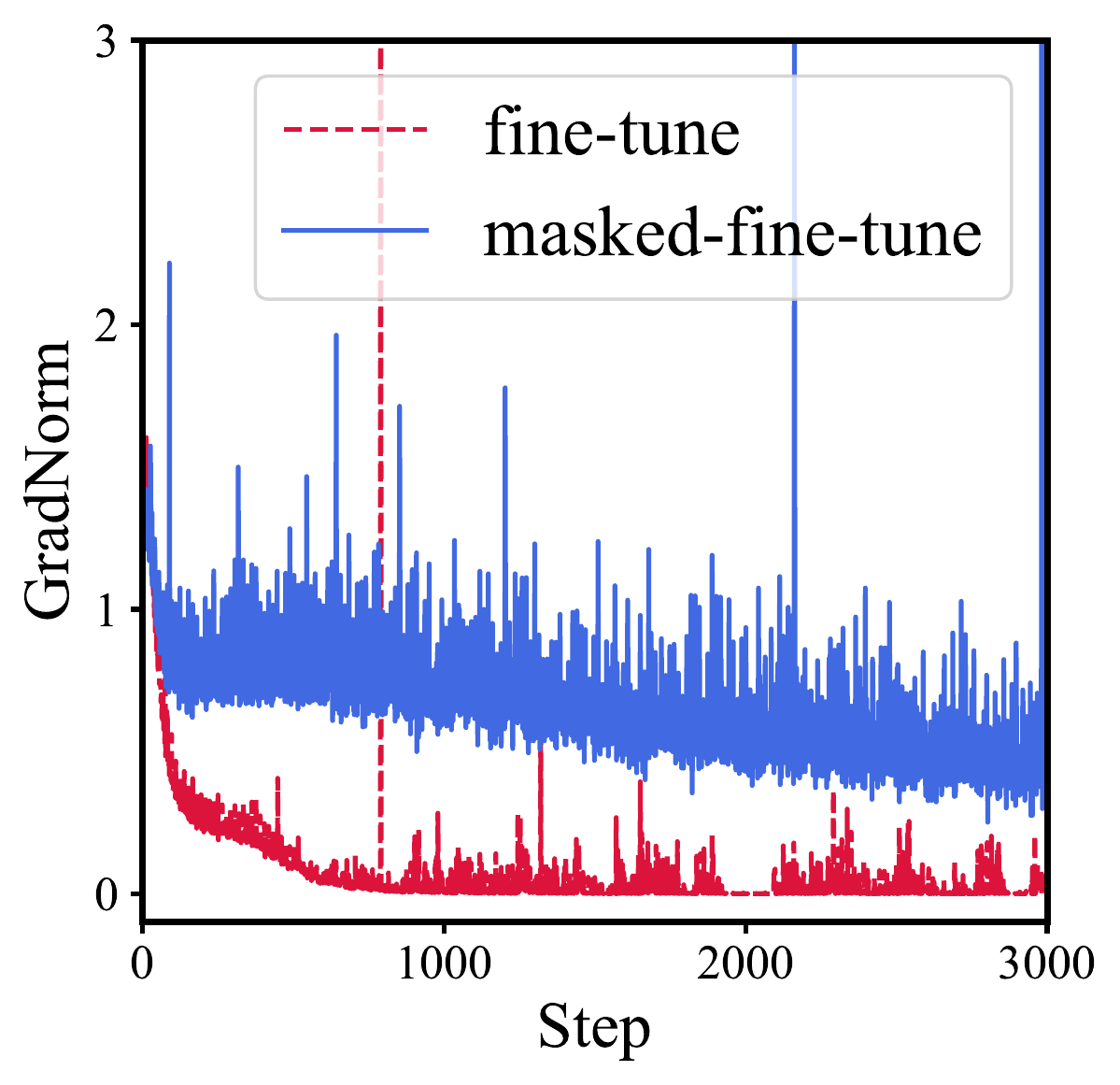}
}
\subfigure[F1]{
\includegraphics[width=0.215\textwidth]{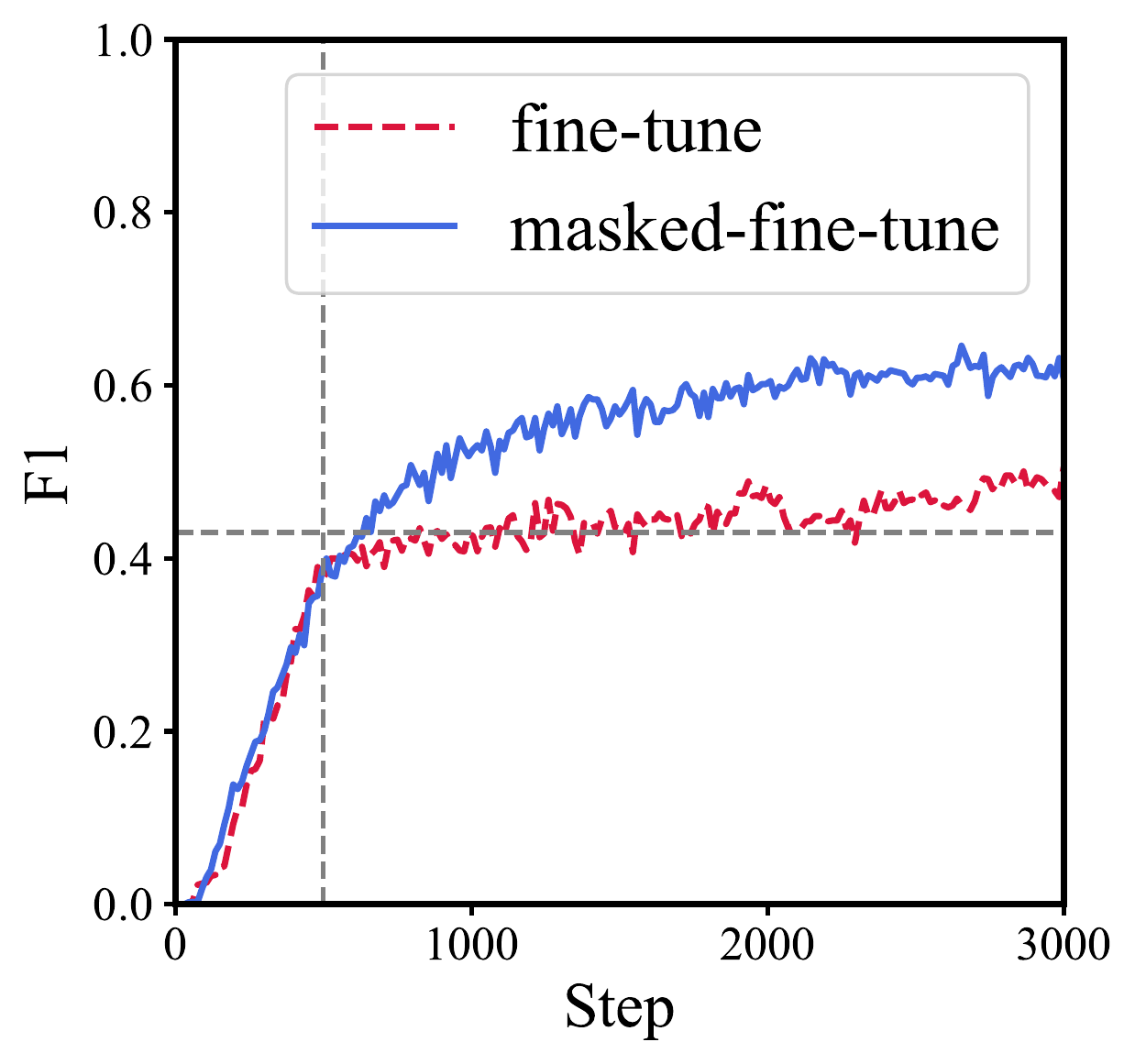}
}
\caption{Gradient and model convergence. In (a), we compute the L2-norm of gradients over all model parameters. In (b), we evaluate the model each 15 steps.}
\label{grad}
\end{figure}

We notice that the error model is relevant to most of the spelling errors, and it is easy to fit the model by memorizing the popular error patterns. As a result, the CSC fine-tuning process converges quickly. We plot the gradient norm curve during training in Figure \ref{grad}. For BERT fine-tuning, the gradient decays quickly. After the gradient norm drops to very small (less than 0.05) in the first few hundreds steps, the F1 score stops increasing. It means that the model has already converged. In contrast, the gradient norm of the Masked-FT model  stays at a high level and the F1 score keeps improving. 

\begin{CJK}{UTF8}{gkai}

Table \ref{singlegrad} reports the gradient norm on each individual token for an example sentence. The gradient produced by BERT fine-tuning is much smaller than that produced by Masked-FT (MFT), indicating that BERT fine-tuning involves less efficient token-level parameter updates across tokens.

\begin{table}[]
\centering
\small
\setlength\tabcolsep{2pt}
\begin{tabular}{@{}l|c|c|c|c|c|c|c|c|c|c@{}}
\toprule
             & 年     & 纪      & 轻     & 就    & 惨     & 遭     & 谢     & 顶      & 。    & \textbf{Sum}  \\ \hline\hline
\textit{FT}  & 0.09   & 0.07   & 0.19   & 0.07  & 0.03  & 0.05   & 0.05   & 0.04   & 0.02  & \textbf{0.79} \\ \hline
\textit{MFT} & 0.27   & 0.10   & 0.40   & 0.19  & 0.53  & 0.68   & 1.16   & 0.92   & 0.26  & \textbf{4.92} \\ \bottomrule
\end{tabular}
\caption{Gradient on each token embedding. We choose a model checkpoint at the early stage of training (two epochs). The sentence is ``(年级 $\rightarrow$  年纪)轻轻就惨遭谢顶。'' (\textit{Shedding of hair at a young (grade$\rightarrow$age).}).}
\label{singlegrad}
\end{table}
\end{CJK}

\section{LEMON Benchmark}

SIGHAN \citep{DBLP:conf/acl-sighan/TsengLCC15} is the current most widely-used benchmark in CSC, but as described in the introduction, it doesn't pose enough challenge to test the generalization ability of CSC models. SIGHAN is exclusively collected from the Chinese essays written by foreign speakers \citep{DBLP:conf/acl-sighan/WuLL13}. That includes 1,100 test examples with a narrow content coverage. Besides, there are 370 distinct edit pairs in the test set, with nearly 70\% overlap with the training set. As a result, a model can achieve a decent score by memorizing the error patterns.

In this paper, we present \textit{LEMON}, a \textit{\textbf{l}arge-scal\textbf{e} \textbf{m}ulti-d\textbf{o}main dataset with \textbf{n}atural spelling errors}, which spans 7 domains, including game (GAM), encyclopedia (ENC), contract (COT), medical care (MEC), car (CAR), novel (NOV), and news (NEW). As opposed to ECSpell \citep{DBLP:journals/corr/abs-2203-10929}, where the typos are deliberately created by human on correct sentences, LEMON consists of over 22,000 examples with natural spelling errors identified from daily human writing, annotated by well-educated native Chinese speakers. The idea is to be as close to the real-life language distribution as possible. LEMON contains 7,627 edit pairs from all domains, which is much more diversified than SIGHAN.

\begin{figure}[t]
\centering
\includegraphics[width=0.42\textwidth]{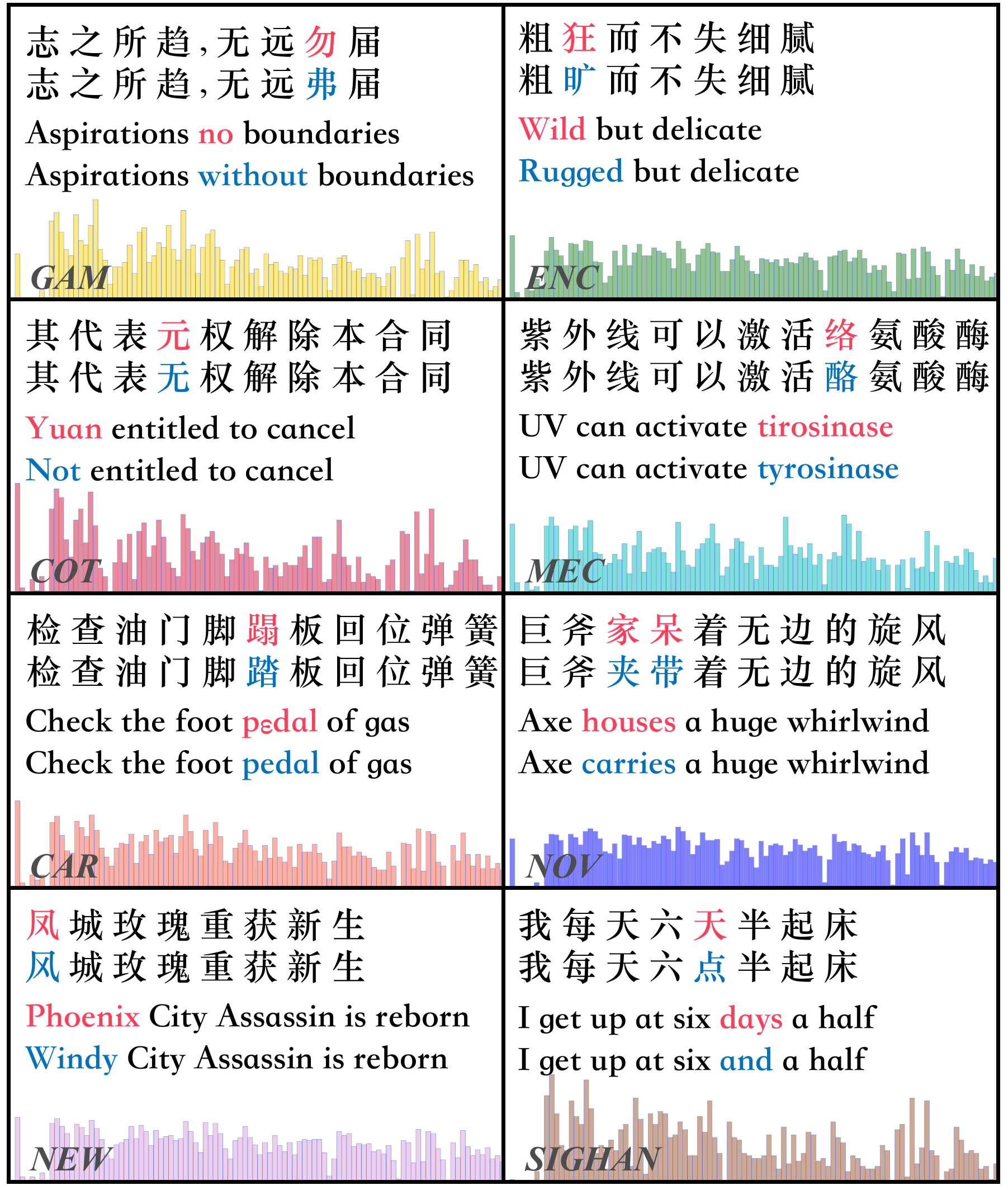}
\caption{A snapshot of LEMON. We also include the SIGHAN-15 test set here for comparison.}
\label{hist}
\end{figure}

Figure \ref{hist} shows some concrete pieces of examples in LEMON. In MEC, for example, we see \textit{tyrosinase} is misspelled, which is a professional word in medicine. The model thus requires certain expertise to correct it. Additionally, the language style of context varies greatly from one domain to another. For example, the expressions in GAM are idiomatic while those in COT are relatively regularized and formal.

The bottom part of each block shows the histogram of all characters in this domain, indicating its lexical distribution. We can see that the lexicon of each domain varies greatly, suggesting different domain-specific language styles. Due to space limitation, further analysis for LEMON is reported in Appendix \ref{appendix:b}.

\section{Masked Fine-Tuning}\label{sec:masked-ft}

The intuition behind masked fine-tuning (Masked-FT) is simple: we want to enhance the learning of language model without perturbing the error model. By equation~\eqref{eqn:bayes}, the language model predicts a token given all other tokens. Thus, we propose to randomly mask a fraction of tokens and train the model to restore them. For training with parallel data, this is equivalent to randomly substituting a fraction of input tokens by a special mask token. The mask token can be any token, as long as it never occurs in an ordinary input. It can be understood as a special ``typo'' that human never makes, thus introducing zero bias to the error model. This technique can be applied to any model architecture.
Empirically, we find that masking 20\% of non-error tokens by \texttt{[MASK]} is the most effective. Other variants, such as using a different masking rate, selecting from both error and non-error tokens, and substituting by \texttt{[unused]}, also works, but they achieve slightly worse results. The ablation study is presented in Section~\ref{sec:further-analysis}.

For training with both parallel (annotated) data and monolingual (unannotated) data, we propose to randomly mask 20\% tokens from the monolingual data, then construct MLM loss~\citep{DBLP:conf/naacl/DevlinCLT19} and add it to the training objective. This is different from generating parallel data by corrupting 20\% tokens. Any corruption rule (e.g. confusion sets) would make assumptions on human errors, thus introduce a bias to the error model. The MLM loss does not introduce any error model bias, and as Section~\ref{sec:empirical-results} shows, it achieves better results in domain transfer.




\section{Empirical Results}\label{sec:empirical-results}

In this section, we compare regular fine-tuning with Masked-FT on a variety of model architectures, and evaluate them on SIGHAN-15, ECSpell, and LEMON. Our implementation is based on \textit{transformers} \citep{wolf-etal-2020-transformers}.

\subsection{Baseline Approaches}

We briefly describe several baseline approaches.

$\bullet$ \textit{BERT}: We fine-tune the BERT model\footnote{\url{https://huggingface.co/bert-base-chinese}}.

$\bullet$ \textit{Soft-Masked BERT}: \citet{DBLP:conf/acl/ZhangHLL20} apply a GRU network as the detector and mask the likely errors in the sequence in a soft way.

$\bullet$ \textit{SpellGCN}: \citet{DBLP:conf/acl/ChengXCJWWCQ20} leverage GCN to integrate phonological and visual features.

$\bullet$ \textit{ConfusBERT}: \citet{DBLP:conf/acl/LiuYYZW20} use the confusion set to guide the mask strategy in MLM pre-training. To idea is to narrow the gap between CSC and MLM.


$\bullet$ \textit{MDCSpell}: \citet{DBLP:conf/acl/ZhuYZM22} design an enhanced detector-corrector network, where two modules are paralleled. The idea is to effectively incorporate the detection clues for decision making.

$\bullet$ \textit{CRASpell}: \citet{DBLP:conf/acl/LiuSYYCYS22} introduce additional errors to the original examples and enhances the local smoothness of the model using KL divergence. The idea is to keep the model robust from noisy context (i.e. with errors).

$\bullet$ \textit{BERT-AT}: \citet{DBLP:conf/acl/LiZZH20} obtain the adversarial examples through character-wise replacement using the confusion set. However, this is time-consuming. As an alternative, we adopt CreAT \citep{wu2023toward}, an end-to-end adversarial training method to obtain the adversarial examples, which perturbs the input embeddings.

We do not take autoregressive models into account in this paper.
It is worth noting that in CSC, BERT-base models remain the primary architecture due to its ability to perform inference for each token in parallel.
It has been shown that in previous work autoregressive models like GPT2 \citep{DBLP:conf/nips/BrownMRSKDNSSAA20} can work much worse on the concerning CSC tasks \citep{DBLP:conf/acl/Li020}.

\subsection{SIGHAN}

\begin{table}[]
\centering
\setlength\tabcolsep{4pt}
\begin{tabular}{l|lll@{}}
\toprule
                                       & Prec.         & Rec.          & F1            \\ \hline\hline
BERT                                   & 73.0          & 72.6          & 72.8          \\
\quad \textit{w/. Masked-FT}           & \textbf{76.7}$_{\uparrow 3.7}$ & \textbf{79.1}$_{\uparrow 6.5}$ & \textbf{77.9}$_{\uparrow 5.1}$ \\
Soft-Masked BERT                       & 67.6          & 72.8          & 70.1          \\
\quad \textit{w/. Masked-FT}           & \textbf{76.3}$_{\uparrow 8.7}$ & \textbf{81.8}$_{\uparrow 9.0}$ & \textbf{79.0}$_{\uparrow 8.9}$ \\ \hdashline\hdashline
MDCSpell$^\dagger$                     & 78.4          & 78.2          & 78.3          \\
SpellGCN$^\dagger$                     & 72.1          & 77.7          & 75.9          \\ \hdashline\hdashline
ConfusBERT$^\dagger$                   & 72.7          & 76.1          & 74.4          \\
DCN$^\dagger$                          & 74.5          & 78.2          & 76.3          \\
PLOME$^\dagger$                        & 75.3          & 79.3          & 77.2          \\
REALISE$^\dagger$                      & 75.9          & 79.9          & 77.8          \\
PHMOSpell$^\dagger$                    & 89.6          & 69.2          & 78.1          \\ \bottomrule
\end{tabular}
\caption{Fine-tuning results on SIGHAN-15. The results in the bottom part requires additional pre-training. $^\dagger$ indicates the result we quote (DCN \citep{DBLP:conf/acl/WangCWWHL21}, PLOME \citep{DBLP:conf/acl/LiuYYZW20}, REALISE \citep{DBLP:conf/acl/XuLZLWCHM21}, PHOMOSpell \citep{DBLP:conf/acl/HuangLJZCWX20}).}
\label{sighan}
\end{table}

SIGHAN-15 \citep{DBLP:conf/acl-sighan/TsengLCC15} is a widely-used benchmark in CSC, which contains 6,476 training examples and 1,100 test examples. We follow the common practice to convert it to simplified Chinese. In addition, we follow the two-stage training setting in most previous work \citep{DBLP:conf/acl/LiuYYZW20,DBLP:conf/acl/ZhuYZM22}, pre-training the model on the public augmented data (271,329 examples) using OCR- and ASR-based generation \citep{DBLP:conf/emnlp/WangSLHZ18}, then in the second stage, training on its own labeled data. We select the best learning rate and batch size in \{1e-5, 2e-5, 5e-5\} and \{32, 128\} respectively for each stage. We train each model for 100,000 steps for the first stage and 10,000 steps for the second.

Table \ref{sighan} summarizes the results on SIGHAN-15. With BERT, Masked-FT achieves very competitive results (improves F1 from 72.8 to 77.9). With Soft-Masked BERT, it achieves the new state-of-the-art on SIGHAN (79.0 F1). Although we have not trained other baseline models with Masked-FT, it is likely that they can get a similar performance boost.

\subsection{ECSpell}

\begin{table}\small
\centering
\setlength\tabcolsep{3.8pt}
\begin{tabular}{@{}l|l|ll|l@{}}
\toprule
                            & Method                              & I-F1          & E-F1          & F1            \\ \hline\hline
\multirow{5}{*}{\small LAW} & vanilla BERT                        & 49.6          & 35.7          & -             \\
                            & BERT                                & 68.4          & 10.0          & 40.2          \\
                            & \quad \textit{w/. Masked-FT} & \textbf{84.9}$_{\uparrow 16.5}$ & \textbf{65.9}$_{\uparrow 55.9}$ & \textbf{76.8}$_{\uparrow 36.6}$ \\
                            & MDCSpell                            & 69.0          & 13.7          & 42.2          \\
                            & \quad \textit{w/. Masked-FT} & \textbf{86.1}$_{\uparrow 17.1}$ & \textbf{73.2}$_{\uparrow 59.5}$ & \textbf{81.1}$_{\uparrow 38.9}$ \\ \hline\hline
\multirow{4}{*}{\small MED} & BERT                                & 35.6          & 5.7           & 26.9          \\
                            & \quad \textit{w/. Masked-FT} & \textbf{46.7}$_{\uparrow 11.1}$ & \textbf{43.2}$_{\uparrow 37.5}$ & \textbf{63.8}$_{\uparrow 36.9}$ \\
                            & MDCSpell                            & 32.1          & 7.4           & 25.7          \\
                            & \quad \textit{w/. Masked-FT} & \textbf{47.9}$_{\uparrow 15.8}$ & \textbf{47.8}$_{\uparrow 40.4}$ & \textbf{72.4}$_{\uparrow 46.7}$ \\ \hline\hline
\multirow{4}{*}{\small ODW} & BERT                                & 54.4          & 7.4           & 26.7          \\
                            & \quad \textit{w/. Masked-FT} & \textbf{71.3}$_{\uparrow 16.9}$ & \textbf{42.4}$_{\uparrow 35}$ & \textbf{62.9}$_{\uparrow 36.2}$ \\
                            & MDCSpell                            & 55.9          & 6.7           & 27.5          \\
                            & \quad \textit{w/. Masked-FT} & \textbf{75.1}$_{\uparrow 19.2}$ & \textbf{51.2}$_{\uparrow 44.5}$ & \textbf{72.0}$_{\uparrow 44.5}$ \\ \bottomrule
\end{tabular}
\caption{Fine-tuning results on ECSpell.}
\label{ecspell}
\end{table}

\begin{table*}[]
\centering
\setlength\tabcolsep{5pt}
\begin{tabular}{@{}llllllllll@{}}
\toprule
                       & \textbf{GAM}          & \textbf{ENC}          & \textbf{COT}          & \textbf{MEC}          & \textbf{CAR}          & \textbf{NOV}          & \textbf{NEW}          & \textbf{SIG}          & Avg           \\ \hline\hline
BERT                   & 27.1                  & 41.6                  & 63.9                  & 47.9                  & 47.6                  & 34.2                  & 50.7                  & 50.6                  & 45.5          \\
\quad \textit{w/. MFT} & 33.3$_{\uparrow 6.2}$ & 45.5$_{\uparrow 3.9}$ & 64.1$_{\uparrow 0.2}$ & 50.9$_{\uparrow 3.0}$ & 52.3$_{\uparrow 4.7}$ & 36.0$_{\uparrow 1.8}$ & 56.0$_{\uparrow 5.3}$ & 53.4$_{\uparrow 2.8}$ & 48.9$_{\uparrow 3.4}$ \\ \hline
Soft-Mased             & 26.3                  & 43.5                  & 63.8                  & 48.8                  & 47.7                  & 34.3                  & 52.7                  & 50.5                  & 45.9          \\
\quad \textit{w/. MFT} & 29.8$_{\uparrow 3.5}$ & 44.6$_{\uparrow 1.1}$ & 65.0$_{\uparrow 1.2}$ & 49.3$_{\uparrow 0.5}$ & 52.0$_{\uparrow 4.3}$ & 37.8$_{\uparrow 3.5}$ & 55.8$_{\uparrow 3.1}$ & 53.4$_{\uparrow 3.0}$ & 48.4$_{\uparrow 2.5}$ \\ \hline
MDCSpell               & 28.2                  & 42.4                  & 63.1                  & 49.4                  & 49.1                  & 35.4                  & 53.9                  & 53.2                  & 46.5          \\
\quad \textit{w/. MFT} & 31.2$_{\uparrow 3.0}$ & 45.9$_{\uparrow 3.5}$ & 65.4$_{\uparrow 2.3}$ & 52.0$_{\uparrow 2.6}$ & 52.6$_{\uparrow 3.5}$ & 38.6$_{\uparrow 3.2}$ & 57.3$_{\uparrow 3.4}$ & 54.7$_{\uparrow 1.5}$ & 49.7$_{\uparrow 3.2}$ \\ \hline
CRASpell               & 22.6                  & 44.5                  & 63.8                  & 48.0                  & 49.6                  & 35.5                  & 53.0                  & 52.4                  & 46.2          \\
\quad \textit{w/. MFT} & 30.7$_{\uparrow 8.1}$ & 48.1$_{\uparrow 3.6}$ & 66.0$_{\uparrow 2.2}$ & 51.7$_{\uparrow 3.7}$ & 51.7$_{\uparrow 2.1}$ & 38.6$_{\uparrow 3.1}$ & 55.9$_{\uparrow 2.9}$ & 55.1$_{\uparrow 2.7}$ & 49.7$_{\uparrow 3.5}$ \\ \hline
BERT-AT                & 25.6                  & 43.0                  & 62.6                  & 49.4                  & 47.5                  & 33.9                  & 51.6                  & 51.0                  & 45.6          \\
\quad \textit{w/. MFT} & 34.4$_{\uparrow 8.8}$ & 47.1$_{\uparrow 4.3}$ & 66.8$_{\uparrow 4.2}$ & 52.0$_{\uparrow 2.6}$ & 51.6$_{\uparrow 4.1}$ & 36.5$_{\uparrow 2.6}$ & 55.0$_{\uparrow 3.4}$ & 53.8$_{\uparrow 2.8}$ & 49.7$_{\uparrow 4.1}$ \\ \bottomrule
\end{tabular}
\caption{Performances on LEMON. We report the F1 scores and also include SIGHAN as the 8$^{\rm th}$ domain (SIG).}
\label{lemon}
\end{table*}

ECSpell \citep{DBLP:journals/corr/abs-2203-10929} is a newly shared CSC dataset with three domains, LAW (1,960 training and 500 test examples), MED (medical treatment, 3,000 training and 500 test) and ODW (official document writing, 1,728 training and 500 test). The hyperparameter search is similar to that in SIGHAN and we train each model for 5,000 steps.

Different form SIGHAN, the test set of ECSpell contains a high proportion ($\approx$ 70\%) of edit pairs that never emerge in the training set. As in Section~\ref{inc-exc}, let EXC be the test subset where the edit pairs are not in the the training set, and INC be the complementary set. We define two new metrics, \textbf{inclusive F1} (I-F1) and \textbf{exclusive F1} (E-F1), to measure the model performance on the two subsets. A higher E-F1 suggests that the model is better at generalizing to unseen errors.

From Table \ref{ecspell}, we see that Masked-FT improves the BERT model's E-F1 by a large scale on all three domains (55.9, 37.5 and 35.0 absolute points). It also generates significant gains on I-F1 (16.5, 11.1 and 16.9 absolute points). This is because that a better language model can assist the error model in making more contextual decisions, even on popular head error patterns. With Masked-FT, BERT and MDCSpell achieve the new state-of-the-art F1 scores on all three domains of ECSpell.

We note that the vanilla BERT performs better than the fine-tuned BERT on E-F1 when the error position is known, but consistently worse than Masked-FT. It means that regular fine-tuning can lead to contextual degeneration, while Masked-FT actually learns a better language model than vanilla BERT.

\subsection{LEMON}

We report two experiments on LEMON. In the first experiment, only monolingual data is used to train the model. We collect monolingual sentences from two general databases \textit{wiki2019zh} and \textit{news2016zh}\footnote{\url{https://github.com/brightmart/nlp_chinese_corpus}} and use the confusion set in \citet{DBLP:conf/acl/LiuYYZW20} to synthesize paired sentences for training. Specifically, we uniformly choose a Chinese character in a sentence and replace it with a counterpart in its confusion set (40\% $\rightarrow$ same pronunciation; 30\% $\rightarrow$ similar pronunciation; 20\% $\rightarrow$ similar glyph; 10\% $\rightarrow$ random). It finally generates 34 million training sentence pairs. We use the same confusion set in the following part, unless otherwise specified.

We select the learning rate in \{1e-5, 2e-5, 5e-5\} and use 8192 as the batch size. Each model is trained for 30,000 steps (more than 7 epochs). We uniformly sample 20\% examples in each domain (no more than 200 examples) and put them together as the development set.

Table \ref{lemon} summarizes the results. We find Masked-FT (shorthand MFT) consistently improves every model and across every domain. It is worth noting that although BERT-AT performs comparably with fine-tuning BERT (only 0.1 gain), the gap grows wider with Masked-FT (0.8 gain). It is known that adversarial training enhances the optimization of the objective function. With regular fine-tuning, it mainly improves error modeling. With Masked-FT, it improves both error modeling and language modeling, resulting in greater performance gains.

In the second experiment, we evaluate on domain transfer. In this setting, we have 2.8M sentence pairs from the news (NEW) domain, annotated by human editors. Our goal is to deploy a model for the medical care (MEC) and the car (CAR) domain. For each of these two domains, we have 10k sentences without any human annotation. We explore two methods to utilize the unannotated data: (1) construct and train with MLM loss, as described in Section~\ref{sec:masked-ft}; (2) generate synthetic data by corrupting unannotated sentences with a confusion set (train with either regular FT or Masked-FT). For both strategies, the model is jointly trained on the 2.8M annotated data along with 10k monolingual data.

From Table \ref{transfer}, we find that incorporating MLM loss on the unannotated data gives higher F1 scores than training with the 2.8M annotated data alone. Furthermore, the MLM loss method works better than the data synthesis method (with or without Mask-FT). We conjecture that the high-quality annotated data has contributed to a precise error model. The additional MLM loss helps learning a better language model for the new domain without changing the error model. On the other hand, the data synthesis method introduces a new error distribution, thus impairs the error model. Overall, the best combination is to jointly train the model on parallel data with Masked-FT, and on monolingual data with MLM loss.

\begin{table}[]
\centering
\small
\setlength\tabcolsep{3pt}
\begin{tabular}{@{}l|l|ccc@{}}
\toprule
Training data                    & Transfer Method & \textbf{NEW}            & \textbf{MEC}            & \textbf{CAR}  \\ \hline
\textit{NEW }                    & -               & 70.7                    & 55.3                    & 64.1          \\ \hdashline
\textit{NEW} + \textit{MEC}      & MLM Loss        & \textbf{72.2}           & \textbf{62.1}           & -             \\
\textit{NEW} + \textit{MEC}      & Synthesis (Masked-FT)      & 71.4                    & 58.1                    & -             \\
\textit{NEW} + \textit{MEC}      & Synthesis (FT)  & 61.4                    & 54.6                    & -             \\ \hline\hline
\textit{NEW} + \textit{CAR}      & MLM Loss        & \textbf{71.1}           & -                       & \textbf{68.4} \\
\textit{NEW} + \textit{CAR}      & Synthesis (Masked-FT)       & 69.4                    & -                       & 65.4          \\
\textit{NEW} + \textit{CAR}      & Synthesis (FT)  & 61.6                    & -                       & 59.5          \\ \bottomrule
\end{tabular}
\caption{Domain transfer results (F1 score). All models are trained with Masked-FT, unless specified as FT, referring to regular fine-tuning.}
\label{transfer}
\end{table}


\section{Further Analysis}\label{sec:further-analysis}

\paragraph{Mask Rate} We investigate the impact from the mask rate $p$. A large $p$ can hurt the training as it wipes out too much contextual information. From Table \ref{p}, we see that the model improves as $p$ goes from 0 to 20\%. Even $p=5\%$ substantially improves E-F1. However, an overly high $p$ can hurt the performance as the context is spoiled.

\begin{table}[]
\centering
\small
\begin{tabular}{@{}l|cccc@{}}
\toprule
Mask rate          & F1            & I-F1          & E-F1          \\ \hline
0\%                & 40.2          & 68.4          & 10.0          \\ \hdashline
5\%                & 62.0          & 73.9          & 47.7          \\
10\%               & 70.1          & 81.3          & 55.2          \\
15\%               & 75.6          & 83.1          & 64.8          \\ \hdashline
20\%               & \textbf{76.8} & \textbf{84.9} & \textbf{65.9} \\
30\%               & 75.7          & 83.2          & 62.3          \\ \hdashline
50\%               & 66.7          & 75.6          & 60.7          \\ \bottomrule
\end{tabular}
\caption{Impact of mask ratio on ECSpell-LAW.}
\label{p}
\end{table}

\paragraph{Mask Strategy} We default to masking the input tokens with the \texttt{[MASK]} token. In fact, any token that does not appear in ordinary inputs can be chosen to perform Masked-FT. From Table \ref{strategy}, we find that masking with \texttt{[unused]} results in similar but slightly lower performance gains. We hypothesize that since \texttt{[MASK]} matches the training of vanilla BERT, it is initialized with a better embedding than that of \texttt{[unused]}. On the other hand, masking with \texttt{[UNK]} leads to a poor result. This is because that \texttt{[UNK]} can occur in ordinary inputs to encode unknown characters. Masking with this token introduces an implicit assumption that when an unknown character appears in the input, it is very likely a spelling error, which is obviously not true. This result highlights the necessity of keeping the error model intact.

Another decision factor is the position to mask. In Table \ref{strategy}, we compare three strategies: masking non-error tokens only, masking error tokens only, and masking any token. We find that the ``masking non-error tokens only'' strategy works the best. This is because that the error model can only be learned from error tokens. Masking error tokens reduces the amount of training data for error modeling, resulting in a slightly worse error model. However, Masked-FT consistently outweighs regular fine-tuning no matter where we mask.

\begin{table}[]
\centering
\small
\begin{tabular}{@{}l|cccl@{}}
\toprule
Mask strategy                 & \textbf{ENC} & \textbf{CAR} & \textbf{NEW} & Avg                        \\ \hline
\textit{fine-tuning}          & 41.6         & 47.6         & 50.7         & 46.6                       \\ \hdashline
\textit{w/. [MASK]}           & 45.5         & 52.3         & 56.0         & \textbf{51.3} ($\uparrow$) \\
\textit{w/. [unused]}         & 44.9         & 52.2         & 55.5         & 50.9 ($\uparrow$)          \\
\textit{w/. [UNK]}            & 39.1         & 45.2         & 47.1         & 43.8 ($\downarrow$)        \\ \hdashline
\textit{mask non-error}       & 45.5         & 52.3         & 56.0         & \textbf{51.3} ($\uparrow$) \\
\textit{mask error}           & 42.9         & 48.2         & 52.2         & 47.8 ($\uparrow$)          \\
\textit{mask any}             & 45.0         & 49.5         & 53.8         & 49.4 ($\uparrow$)          \\\bottomrule
\end{tabular}
\caption{Comparison of mask strategies on three LEMON domains (F1 score). The mask rate is 0.2.}
\label{strategy}
\end{table}

\paragraph{vs. Data Augmentation via Confusion Set} A popular data augmentation strategy is to randomly substitute a certain fraction of tokens with a misspelled token from the confusion set. \citet{DBLP:conf/acl/LiuYYZW20} use the confusion set to guide the masking strategy in MLM pre-training. We apply the same confusion set substitution rules to fine-tuning. As shown in Table \ref{confusion-sig}, using a confusion set for data augmentation helps in the pre-training stage, but it does not help in the fine-tuning stage. Again, this is due to the fact that any confusion set introduces a bias to the error model. In particular, the confusion set substitution injects large amount of errors that humans would not make in practice. As a result, the model will learn to detect and correct errors in an overly aggressive manner. 

\begin{table}[H]
\centering
\small
\begin{tabular}{@{}l|l|ccc@{}}
\toprule
                        & Method                              & Prec.                   & Rec.                    & F1                      \\ \hline
\multirow{3}{*}{SIGHAN} & \textit{Masked-FT}           & \textbf{76.7}           & \textbf{79.1}           & \textbf{77.9}           \\
                        & \textit{confusion-FT}           & 63.9                    & 75.2                    & 69.1                    \\
                        & \textit{confusion-pretrain}$^\dagger$ & 72.7                    & 76.1                    & 74.4                    \\ \bottomrule
\end{tabular}
\caption{Masked-FT vs. confusion set (F1 score).}
\label{confusion-sig}
\end{table}
\vspace{-20pt}
\begin{table}[H]
\centering
\small
\begin{tabular}{@{}l|l|ccc@{}}
\toprule
                        & Method                              & \textbf{ENC}            & \textbf{CAR}            & \textbf{NEW}            \\ \hline
\multirow{3}{*}{LEMON}  & \textit{Masked-FT}           & \textbf{45.5}           & \textbf{52.3}           & \textbf{56.0}           \\
                        & \textit{confusion-FT}           & 35.2                    & 43.4                    & 46.3                    \\
                        & \textit{mixed-FT}            & 40.7                    & 47.4                    & 50.5                    \\ \bottomrule
\end{tabular}
\caption{Masked-FT vs. confusion set (F1 score).}
\label{confusion-lem}
\end{table}

Table~\ref{confusion-lem} reports a similar comparison on LEMON. Again, Masked-FT consistently outperforms fine-tuning with confusion set substitution. We also compare with the ``mixed'' strategy proposed by \cite{DBLP:conf/aaai/ZhaoW20}: with 50\% probability, masking the sentence, and with the remaining 50\% probability, corrupting the sentence via the confusion set. The result of the ``mixed'' strategy interpolates between the two extremes, suggesting that a mixing strategy cannot offset the error model bias caused by the confusion set.

\begin{CJK}{UTF8}{gkai}
\paragraph{Case Study} We study two concrete examples in Table \ref{case} where CSC is context dependent.
For the first case (\textit{It seems no one has ever {\color[HTML]{DC143C}found out} silver taels.}),
the fine-tuned model wants to correct \textit{found out} to be \textit{took out}, while Mask-FT does not make any change. Both \textit{found out silver taels} and \textit{took out silver taels} are reasonable combinations. According to the context, however, we can reason that someone is digging for treasure. Hence, \textit{found out silver taels} is more appropriate. For the second case (\textit{There was a smart person who applied for a job with a salary of 1 yuan for the first year, 2 {\color[HTML]{DC143C}years} ($\rightarrow$ {\color[HTML]{00008B}yuan}) for the second...}), we can reason the second \textit{year} should be corrected to \textit{yuan} because the previous context mentions \textit{salary}, while the fine-tuned model is not able to do so.

\begin{table}[]
\centering
\small
\setlength\tabcolsep{4pt}
\begin{tabular}{@{}ll@{}}
\toprule
\textit{source}& 但好像从没见人{\color[HTML]{DC143C}淘}出过银两。 \\
\textit{target}& 但好像从没见人{\color[HTML]{DC143C}淘}出过银两。 \\
\textit{FT}& 但好像从没见人{\color[HTML]{00008B}掏}出过银两。 \\
\textit{Masked-FT}& 但好像从没见人{\color[HTML]{DC143C}淘}出过银两。 \\ \hline\hline
\textit{source}& 有一聪明人应聘年薪只要1元,第二年2{\color[HTML]{DC143C}年}... \\
\textit{target}& 有一聪明人应聘年薪只要1元,第二年2{\color[HTML]{00008B}元}... \\
\textit{FT}& 有一聪明人应聘年薪只要1元,第二年2{\color[HTML]{DC143C}年}... \\
\textit{Masked-FT}& 有一聪明人应聘年薪只要1元,第二年2{\color[HTML]{00008B}元}... \\ \bottomrule
\end{tabular}
\caption{Case study selected from LEMON.}
\label{case}
\end{table}
\end{CJK}

\begin{CJK}{UTF8}{gkai}
\paragraph{Error analysis} Though Masked-FT exhibits powerful potential, we further study its error cases to enlighten future research.
We illustrate two typical error cases in Table \ref{error}.
For the first case, ``洛汀新'' (\textit{Lotensin}) is a particular kind of pill, while Mask-FT cannot allow the model to acquire professional knowledge.
It suggests that a universal correction system necessitates domain-specific data or knowledge for stronger adaption to some domain like medicine, science, with a wide range of expertise.
For the second case, the model wrongly corrects ``妆'' (\textit{makeup}) to ``浴'' (\textit{bathing}) because of the subsequent context ``保持皮肤洁净'' (\textit{keep skin clean}).
It implies a subtle trade-off between language model and error model. Of course, this is an extreme case, which rarely occurs.

\begin{table}[]
\centering
\small
\setlength\tabcolsep{4pt}
\begin{tabular}{@{}ll@{}}
\toprule
\textit{source}& 可以换成洛{\color[HTML]{DC143C}听}新，一天一片... \\
\textit{target}& 可以换成洛{\color[HTML]{00008B}汀}新，一天一片... \\
\textit{Masked-FT}& 可以换成洛{\color[HTML]{DC143C}听}新，一天一片... \\ \hline\hline
\textit{source}& 不要随便使用化{\color[HTML]{DC143C}妆}品，保持皮肤洁净... \\
\textit{target}& 不要随便使用化{\color[HTML]{DC143C}妆}品，保持皮肤洁净... \\
\textit{Masked-FT}& 不要随便使用化{\color[HTML]{00008B}浴}品，保持皮肤洁净... \\ \bottomrule
\end{tabular}
\caption{Error analysis selected from ECSpell-MED.}
\label{error}
\end{table}
\end{CJK}

\section{Related Work}

For Chinese spelling correction, BERT \citep{DBLP:conf/naacl/DevlinCLT19,DBLP:journals/corr/abs-1907-11692,DBLP:conf/emnlp/CuiC000H20} is the straightforward backbone model. There is a line of work on improving the model architecture on top of BERT, such as imposing masking signals to those potential error tokens to improve error detection \citep{DBLP:conf/acl/ZhangHLL20}, incorporating multi-modal knowledge (e.g. pronunciation, glyph) \citep{DBLP:conf/acl/ChengXCJWWCQ20,DBLP:conf/acl/LiuYYZW20,DBLP:conf/acl/HuangLJZCWX20,DBLP:conf/acl/XuLZLWCHM21,DBLP:conf/acl/ZhangPZWHSWW21},
using multi-task network to explicitly let the model detect \citep{DBLP:conf/acl/ZhuYZM22} or predict the pronunciation \citep{DBLP:conf/acl/LiuYYZW20}.
Another major category is data augmentation, with the goal of synthesizing efficient training data. Existing data augmentation techniques are based on homophone substitution, random substitution or confusion sets \citep{DBLP:conf/emnlp/WangSLHZ18,DBLP:conf/acl/WangTZ19,DBLP:conf/acl/LiuYYZW20,DBLP:conf/acl/GuoNWZX21}.

The decomposition of CSC into a language model and an error model is inspired by the classical noisy channel theory~\citep{DBLP:conf/coling/KernighanCG90}. The masked-FT method proposed in this paper is similar to the ``dynamic masking'' method proposed by ~\citet{DBLP:conf/aaai/ZhaoW20}. However, there are a few differences between the two studies. First, \citet{DBLP:conf/aaai/ZhaoW20} describes dynamic masking as a data augmentation method, and proposes to mix it with other data augmentation techniques such as confusion set substitution; in contrast, we describe masked-FT as a mean to enhance language modeling without perturbing error modeling, demonstrating both theoretically and empirically that it should be carried out alone without mixing with data augmentation. Second, we study domain transfer with monolingual data, showing that MLM training performs better than training with synthesized data. Again, it verifies our language/error decomposition theory and to the best of our knowledge, was not discussed in previous work.

\section{Conclusion}

This paper presents qualitative analysis and shows that existing CSC models lean to over-fit the error model and under-fit the language model. A simple yet effective method is thus presented to encourage a better language model learning. Empirical results demonstrate that the simple method achieves new state-of-the-art results on public benchmarks, including on LENON, a new large-scale challenging benchmark released with this paper.

\section*{Limitations}

We have not tested all possible recent methods on LEMON. We have used expensive GPU resources to speed up the training process on LEMON, with 8 NVIDIA A100 sheets, but consistent results can also be obtained with 8 V100 sheets. Our work focuses on Chinese. Other languages, such as Japanese and Korean, could benefit from the same technique, but have not been studied in this work.


\bibliography{anthology,custom}

\begin{thebibliography}{29}
\expandafter\ifx\csname natexlab\endcsname\relax\def\natexlab#1{#1}\fi

\bibitem[{Afli et~al.(2016)Afli, Qiu, Way, and
  Sheridan}]{DBLP:conf/lrec/AfliQWS16}
Haithem Afli, Zhengwei Qiu, Andy Way, and P{\'{a}}raic Sheridan. 2016.
\newblock \href
  {http://www.lrec-conf.org/proceedings/lrec2016/summaries/280.html} {Using
  {SMT} for {OCR} error correction of historical texts}.
\newblock In \emph{Proceedings of the Tenth International Conference on
  Language Resources and Evaluation {LREC} 2016, Portoro{\v{z}}, Slovenia, May
  23-28, 2016}. European Language Resources Association {(ELRA)}.

\bibitem[{Brown et~al.(2020)Brown, Mann, Ryder, Subbiah, Kaplan, Dhariwal,
  Neelakantan, Shyam, Sastry, Askell, Agarwal, Herbert{-}Voss, Krueger,
  Henighan, Child, Ramesh, Ziegler, Wu, Winter, Hesse, Chen, Sigler, Litwin,
  Gray, Chess, Clark, Berner, McCandlish, Radford, Sutskever, and
  Amodei}]{DBLP:conf/nips/BrownMRSKDNSSAA20}
Tom~B. Brown, Benjamin Mann, Nick Ryder, Melanie Subbiah, Jared Kaplan,
  Prafulla Dhariwal, Arvind Neelakantan, Pranav Shyam, Girish Sastry, Amanda
  Askell, Sandhini Agarwal, Ariel Herbert{-}Voss, Gretchen Krueger, Tom
  Henighan, Rewon Child, Aditya Ramesh, Daniel~M. Ziegler, Jeffrey Wu, Clemens
  Winter, Christopher Hesse, Mark Chen, Eric Sigler, Mateusz Litwin, Scott
  Gray, Benjamin Chess, Jack Clark, Christopher Berner, Sam McCandlish, Alec
  Radford, Ilya Sutskever, and Dario Amodei. 2020.
\newblock \href
  {https://proceedings.neurips.cc/paper/2020/hash/1457c0d6bfcb4967418bfb8ac142f64a-Abstract.html}
  {Language models are few-shot learners}.
\newblock In \emph{Advances in Neural Information Processing Systems 33: Annual
  Conference on Neural Information Processing Systems 2020, NeurIPS 2020,
  December 6-12, 2020, virtual}.

\bibitem[{Cheng et~al.(2020)Cheng, Xu, Chen, Jiang, Wang, Wang, Chu, and
  Qi}]{DBLP:conf/acl/ChengXCJWWCQ20}
Xingyi Cheng, Weidi Xu, Kunlong Chen, Shaohua Jiang, Feng Wang, Taifeng Wang,
  Wei Chu, and Yuan Qi. 2020.
\newblock \href {https://doi.org/10.18653/v1/2020.acl-main.81} {Spellgcn:
  Incorporating phonological and visual similarities into language models for
  chinese spelling check}.
\newblock In \emph{Proceedings of the 58th Annual Meeting of the Association
  for Computational Linguistics, {ACL} 2020, Online, July 5-10, 2020}, pages
  871--881. Association for Computational Linguistics.

\bibitem[{Cui et~al.(2020)Cui, Che, Liu, Qin, Wang, and
  Hu}]{DBLP:conf/emnlp/CuiC000H20}
Yiming Cui, Wanxiang Che, Ting Liu, Bing Qin, Shijin Wang, and Guoping Hu.
  2020.
\newblock \href {https://doi.org/10.18653/v1/2020.findings-emnlp.58}
  {Revisiting pre-trained models for chinese natural language processing}.
\newblock In \emph{Findings of the Association for Computational Linguistics:
  {EMNLP} 2020, Online Event, 16-20 November 2020}, volume {EMNLP} 2020 of
  \emph{Findings of {ACL}}, pages 657--668. Association for Computational
  Linguistics.

\bibitem[{Devlin et~al.(2019)Devlin, Chang, Lee, and
  Toutanova}]{DBLP:conf/naacl/DevlinCLT19}
Jacob Devlin, Ming{-}Wei Chang, Kenton Lee, and Kristina Toutanova. 2019.
\newblock \href {https://doi.org/10.18653/v1/n19-1423} {{BERT:} pre-training of
  deep bidirectional transformers for language understanding}.
\newblock In \emph{Proceedings of the 2019 Conference of the North American
  Chapter of the Association for Computational Linguistics: Human Language
  Technologies, {NAACL-HLT} 2019, Minneapolis, MN, USA, June 2-7, 2019, Volume
  1 (Long and Short Papers)}, pages 4171--4186. Association for Computational
  Linguistics.

\bibitem[{Gao et~al.(2010)Gao, Li, Micol, Quirk, and
  Sun}]{DBLP:conf/coling/GaoLMQS10}
Jianfeng Gao, Xiaolong Li, Daniel Micol, Chris Quirk, and Xu~Sun. 2010.
\newblock \href {https://aclanthology.org/C10-1041/} {A large scale
  ranker-based system for search query spelling correction}.
\newblock In \emph{{COLING} 2010, 23rd International Conference on
  Computational Linguistics, Proceedings of the Conference, 23-27 August 2010,
  Beijing, China}, pages 358--366. Tsinghua University Press.

\bibitem[{Guo et~al.(2021)Guo, Ni, Wang, Zhu, and
  Xie}]{DBLP:conf/acl/GuoNWZX21}
Zhao Guo, Yuan Ni, Keqiang Wang, Wei Zhu, and Guotong Xie. 2021.
\newblock \href {https://doi.org/10.18653/v1/2021.findings-acl.122} {Global
  attention decoder for chinese spelling error correction}.
\newblock In \emph{Findings of the Association for Computational Linguistics:
  {ACL/IJCNLP} 2021, Online Event, August 1-6, 2021}, volume {ACL/IJCNLP} 2021
  of \emph{Findings of {ACL}}, pages 1419--1428. Association for Computational
  Linguistics.

\bibitem[{Gupta et~al.(2021)Gupta, Corro, Broscheit, Hoffart, and
  Brenner}]{DBLP:conf/emnlp/GuptaCBHB21}
Harsh Gupta, Luciano~Del Corro, Samuel Broscheit, Johannes Hoffart, and Eliot
  Brenner. 2021.
\newblock \href {https://doi.org/10.18653/v1/2021.emnlp-main.680} {Unsupervised
  multi-view post-ocr error correction with language models}.
\newblock In \emph{Proceedings of the 2021 Conference on Empirical Methods in
  Natural Language Processing, {EMNLP} 2021, Virtual Event / Punta Cana,
  Dominican Republic, 7-11 November, 2021}, pages 8647--8652. Association for
  Computational Linguistics.

\bibitem[{Huang et~al.(2021)Huang, Li, Jiang, Zhang, Chen, Wang, and
  Xiao}]{DBLP:conf/acl/HuangLJZCWX20}
Li~Huang, Junjie Li, Weiwei Jiang, Zhiyu Zhang, Minchuan Chen, Shaojun Wang,
  and Jing Xiao. 2021.
\newblock \href {https://doi.org/10.18653/v1/2021.acl-long.464} {Phmospell:
  Phonological and morphological knowledge guided chinese spelling check}.
\newblock In \emph{Proceedings of the 59th Annual Meeting of the Association
  for Computational Linguistics and the 11th International Joint Conference on
  Natural Language Processing, {ACL/IJCNLP} 2021, (Volume 1: Long Papers),
  Virtual Event, August 1-6, 2021}, pages 5958--5967. Association for
  Computational Linguistics.

\bibitem[{Kernighan et~al.(1990)Kernighan, Church, and
  Gale}]{DBLP:conf/coling/KernighanCG90}
Mark~D. Kernighan, Kenneth~Ward Church, and William~A. Gale. 1990.
\newblock \href {https://aclanthology.org/C90-2036/} {A spelling correction
  program based on a noisy channel model}.
\newblock In \emph{13th International Conference on Computational Linguistics,
  {COLING} 1990, University of Helsinki, Finland, August 20-25, 1990}, pages
  205--210.

\bibitem[{Li et~al.(2021)Li, Zhang, Zheng, and Huang}]{DBLP:conf/acl/LiZZH20}
Chong Li, Cenyuan Zhang, Xiaoqing Zheng, and Xuanjing Huang. 2021.
\newblock \href {https://doi.org/10.18653/v1/2021.acl-short.56} {Exploration
  and exploitation: Two ways to improve chinese spelling correction models}.
\newblock In \emph{Proceedings of the 59th Annual Meeting of the Association
  for Computational Linguistics and the 11th International Joint Conference on
  Natural Language Processing, {ACL/IJCNLP} 2021, (Volume 2: Short Papers),
  Virtual Event, August 1-6, 2021}, pages 441--446. Association for
  Computational Linguistics.

\bibitem[{Li and Shi(2021)}]{DBLP:conf/acl/Li020}
Piji Li and Shuming Shi. 2021.
\newblock \href {https://doi.org/10.18653/v1/2021.acl-long.385} {Tail-to-tail
  non-autoregressive sequence prediction for chinese grammatical error
  correction}.
\newblock In \emph{Proceedings of the 59th Annual Meeting of the Association
  for Computational Linguistics and the 11th International Joint Conference on
  Natural Language Processing, {ACL/IJCNLP} 2021, (Volume 1: Long Papers),
  Virtual Event, August 1-6, 2021}, pages 4973--4984. Association for
  Computational Linguistics.

\bibitem[{Liu et~al.(2022)Liu, Song, Yue, Yang, Cai, Yu, and
  Sun}]{DBLP:conf/acl/LiuSYYCYS22}
Shulin Liu, Shengkang Song, Tianchi Yue, Tao Yang, Huihui Cai, Tinghao Yu, and
  Shengli Sun. 2022.
\newblock \href {https://doi.org/10.18653/v1/2022.findings-acl.237} {Craspell:
  {A} contextual typo robust approach to improve chinese spelling correction}.
\newblock In \emph{Findings of the Association for Computational Linguistics:
  {ACL} 2022, Dublin, Ireland, May 22-27, 2022}, pages 3008--3018. Association
  for Computational Linguistics.

\bibitem[{Liu et~al.(2021)Liu, Yang, Yue, Zhang, and
  Wang}]{DBLP:conf/acl/LiuYYZW20}
Shulin Liu, Tao Yang, Tianchi Yue, Feng Zhang, and Di~Wang. 2021.
\newblock \href {https://doi.org/10.18653/v1/2021.acl-long.233} {{PLOME:}
  pre-training with misspelled knowledge for chinese spelling correction}.
\newblock In \emph{Proceedings of the 59th Annual Meeting of the Association
  for Computational Linguistics and the 11th International Joint Conference on
  Natural Language Processing, {ACL/IJCNLP} 2021, (Volume 1: Long Papers),
  Virtual Event, August 1-6, 2021}, pages 2991--3000. Association for
  Computational Linguistics.

\bibitem[{Liu et~al.(2019)Liu, Ott, Goyal, Du, Joshi, Chen, Levy, Lewis,
  Zettlemoyer, and Stoyanov}]{DBLP:journals/corr/abs-1907-11692}
Yinhan Liu, Myle Ott, Naman Goyal, Jingfei Du, Mandar Joshi, Danqi Chen, Omer
  Levy, Mike Lewis, Luke Zettlemoyer, and Veselin Stoyanov. 2019.
\newblock \href {http://arxiv.org/abs/1907.11692} {Roberta: {A} robustly
  optimized {BERT} pretraining approach}.
\newblock \emph{CoRR}, abs/1907.11692.

\bibitem[{Lv et~al.(2022)Lv, Cao, Geng, Ai, Yan, and
  Fu}]{DBLP:journals/corr/abs-2203-10929}
Qi~Lv, Ziqiang Cao, Lei Geng, Chunhui Ai, Xu~Yan, and Guohong Fu. 2022.
\newblock \href {https://doi.org/10.48550/arXiv.2203.10929} {General and domain
  adaptive chinese spelling check with error consistent pretraining}.
\newblock \emph{CoRR}, abs/2203.10929.

\bibitem[{Martins and Silva(2004)}]{DBLP:conf/tal/MartinsS04}
Bruno Martins and M{\'{a}}rio~J. Silva. 2004.
\newblock \href {https://doi.org/10.1007/978-3-540-30228-5\_33} {Spelling
  correction for search engine queries}.
\newblock In \emph{Advances in Natural Language Processing, 4th International
  Conference, EsTAL 2004, Alicante, Spain, October 20-22, 2004, Proceedings},
  volume 3230 of \emph{Lecture Notes in Computer Science}, pages 372--383.
  Springer.

\bibitem[{Tseng et~al.(2015)Tseng, Lee, Chang, and
  Chen}]{DBLP:conf/acl-sighan/TsengLCC15}
Yuen{-}Hsien Tseng, Lung{-}Hao Lee, Li{-}Ping Chang, and Hsin{-}Hsi Chen. 2015.
\newblock \href {https://doi.org/10.18653/v1/W15-3106} {Introduction to
  {SIGHAN} 2015 bake-off for chinese spelling check}.
\newblock In \emph{Proceedings of the Eighth {SIGHAN} Workshop on Chinese
  Language Processing, SIGHAN@IJCNLP 2015, Beijing, China, July 30-31, 2015},
  pages 32--37. Association for Computational Linguistics.

\bibitem[{Wang et~al.(2021)Wang, Che, Wu, Wang, Hu, and
  Liu}]{DBLP:conf/acl/WangCWWHL21}
Baoxin Wang, Wanxiang Che, Dayong Wu, Shijin Wang, Guoping Hu, and Ting Liu.
  2021.
\newblock \href {https://doi.org/10.18653/v1/2021.findings-acl.216} {Dynamic
  connected networks for chinese spelling check}.
\newblock In \emph{Findings of the Association for Computational Linguistics:
  {ACL/IJCNLP} 2021, Online Event, August 1-6, 2021}, volume {ACL/IJCNLP} 2021
  of \emph{Findings of {ACL}}, pages 2437--2446. Association for Computational
  Linguistics.

\bibitem[{Wang et~al.(2018)Wang, Song, Li, Han, and
  Zhang}]{DBLP:conf/emnlp/WangSLHZ18}
Dingmin Wang, Yan Song, Jing Li, Jialong Han, and Haisong Zhang. 2018.
\newblock \href {https://doi.org/10.18653/v1/d18-1273} {A hybrid approach to
  automatic corpus generation for chinese spelling check}.
\newblock In \emph{Proceedings of the 2018 Conference on Empirical Methods in
  Natural Language Processing, Brussels, Belgium, October 31 - November 4,
  2018}, pages 2517--2527. Association for Computational Linguistics.

\bibitem[{Wang et~al.(2019)Wang, Tay, and Zhong}]{DBLP:conf/acl/WangTZ19}
Dingmin Wang, Yi~Tay, and Li~Zhong. 2019.
\newblock \href {https://doi.org/10.18653/v1/p19-1578} {Confusionset-guided
  pointer networks for chinese spelling check}.
\newblock In \emph{Proceedings of the 57th Conference of the Association for
  Computational Linguistics, {ACL} 2019, Florence, Italy, July 28- August 2,
  2019, Volume 1: Long Papers}, pages 5780--5785. Association for Computational
  Linguistics.

\bibitem[{Wolf et~al.(2020)Wolf, Debut, Sanh, Chaumond, Delangue, Moi, Cistac,
  Rault, Louf, Funtowicz, Davison, Shleifer, von Platen, Ma, Jernite, Plu, Xu,
  Le~Scao, Gugger, Drame, Lhoest, and Rush}]{wolf-etal-2020-transformers}
Thomas Wolf, Lysandre Debut, Victor Sanh, Julien Chaumond, Clement Delangue,
  Anthony Moi, Pierric Cistac, Tim Rault, Remi Louf, Morgan Funtowicz, Joe
  Davison, Sam Shleifer, Patrick von Platen, Clara Ma, Yacine Jernite, Julien
  Plu, Canwen Xu, Teven Le~Scao, Sylvain Gugger, Mariama Drame, Quentin Lhoest,
  and Alexander Rush. 2020.
\newblock \href {https://doi.org/10.18653/v1/2020.emnlp-demos.6} {Transformers:
  State-of-the-art natural language processing}.
\newblock In \emph{Proceedings of the 2020 Conference on Empirical Methods in
  Natural Language Processing: System Demonstrations}, pages 38--45, Online.
  Association for Computational Linguistics.

\bibitem[{Wu et~al.(2023)Wu, Liu, Shi, hai zhao, and Zhang}]{wu2023toward}
Hongqiu Wu, Yongxiang Liu, Hanwen Shi, hai zhao, and Min Zhang. 2023.
\newblock \href {https://openreview.net/forum?id=xZD10GhCvM} {Toward
  adversarial training on contextualized language representation}.
\newblock In \emph{The Eleventh International Conference on Learning
  Representations}.

\bibitem[{Wu et~al.(2013)Wu, Liu, and Lee}]{DBLP:conf/acl-sighan/WuLL13}
Shih{-}Hung Wu, Chao{-}Lin Liu, and Lung{-}Hao Lee. 2013.
\newblock \href {https://aclanthology.org/W13-4406/} {Chinese spelling check
  evaluation at {SIGHAN} bake-off 2013}.
\newblock In \emph{Proceedings of the Seventh {SIGHAN} Workshop on Chinese
  Language Processing, SIGHAN@IJCNLP 2013, Nagoya, Japan, October 14-18, 2013},
  pages 35--42. Asian Federation of Natural Language Processing.

\bibitem[{Xu et~al.(2021)Xu, Li, Zhou, Li, Wang, Cao, Huang, and
  Mao}]{DBLP:conf/acl/XuLZLWCHM21}
Heng{-}Da Xu, Zhongli Li, Qingyu Zhou, Chao Li, Zizhen Wang, Yunbo Cao, Heyan
  Huang, and Xian{-}Ling Mao. 2021.
\newblock \href {https://doi.org/10.18653/v1/2021.findings-acl.64} {Read,
  listen, and see: Leveraging multimodal information helps chinese spell
  checking}.
\newblock In \emph{Findings of the Association for Computational Linguistics:
  {ACL/IJCNLP} 2021, Online Event, August 1-6, 2021}, volume {ACL/IJCNLP} 2021
  of \emph{Findings of {ACL}}, pages 716--728. Association for Computational
  Linguistics.

\bibitem[{Zhang et~al.(2021)Zhang, Pang, Zhang, Wang, He, Sun, Wu, and
  Wang}]{DBLP:conf/acl/ZhangPZWHSWW21}
Ruiqing Zhang, Chao Pang, Chuanqiang Zhang, Shuohuan Wang, Zhongjun He, Yu~Sun,
  Hua Wu, and Haifeng Wang. 2021.
\newblock \href {https://doi.org/10.18653/v1/2021.findings-acl.198} {Correcting
  chinese spelling errors with phonetic pre-training}.
\newblock In \emph{Findings of the Association for Computational Linguistics:
  {ACL/IJCNLP} 2021, Online Event, August 1-6, 2021}, volume {ACL/IJCNLP} 2021
  of \emph{Findings of {ACL}}, pages 2250--2261. Association for Computational
  Linguistics.

\bibitem[{Zhang et~al.(2020)Zhang, Huang, Liu, and
  Li}]{DBLP:conf/acl/ZhangHLL20}
Shaohua Zhang, Haoran Huang, Jicong Liu, and Hang Li. 2020.
\newblock \href {https://doi.org/10.18653/v1/2020.acl-main.82} {Spelling error
  correction with soft-masked {BERT}}.
\newblock In \emph{Proceedings of the 58th Annual Meeting of the Association
  for Computational Linguistics, {ACL} 2020, Online, July 5-10, 2020}, pages
  882--890. Association for Computational Linguistics.

\bibitem[{Zhao and Wang(2020)}]{DBLP:conf/aaai/ZhaoW20}
Zewei Zhao and Houfeng Wang. 2020.
\newblock \href {https://ojs.aaai.org/index.php/AAAI/article/view/5476}
  {Maskgec: Improving neural grammatical error correction via dynamic masking}.
\newblock In \emph{The Thirty-Fourth {AAAI} Conference on Artificial
  Intelligence, {AAAI} 2020, The Thirty-Second Innovative Applications of
  Artificial Intelligence Conference, {IAAI} 2020, The Tenth {AAAI} Symposium
  on Educational Advances in Artificial Intelligence, {EAAI} 2020, New York,
  NY, USA, February 7-12, 2020}, pages 1226--1233. {AAAI} Press.

\bibitem[{Zhu et~al.(2022)Zhu, Ying, Zhang, and Mao}]{DBLP:conf/acl/ZhuYZM22}
Chenxi Zhu, Ziqiang Ying, Boyu Zhang, and Feng Mao. 2022.
\newblock \href {https://doi.org/10.18653/v1/2022.findings-acl.98} {Mdcspell:
  {A} multi-task detector-corrector framework for chinese spelling correction}.
\newblock In \emph{Findings of the Association for Computational Linguistics:
  {ACL} 2022, Dublin, Ireland, May 22-27, 2022}, pages 1244--1253. Association
  for Computational Linguistics.

\end{thebibliography}
\bibliographystyle{acl_natbib}

\appendix

\section{Derivation of Equation~\eqref{eqn:bayes}}
\label{appendix:a}

Let the input sentence be $X = (x_1,...,x_n)$ and output sentence be $Y = (y_1,...,y_n)$. Given $X$, the BERT model predicts each element of $Y$ separately, namely computing $P(y_i|X)$ for $i=1,2,...,n$. Let $x_{-i} = (x_1,...,x_{i-1},x_{i+1},...,x_n)$, then $P(y_i|X) = P(y_i|x_i, x_{-i})$. By Bayes Rule:
\[
P(y_i|x_i, x_{-i}) = \frac{P(y_i|x_{-i})P(x_i|y_i,x_{-i})}{P(x_i|x_{-i})}
\]
Notice that $P(x_i|x_{-i})$ is a constant for varying $y_i$, thus the left-hand side is proportional to the numerator, namely
\[
P(y_i|x_i, x_{-i}) \propto P(y_i|x_{-i})P(x_i|y_i,x_{-i}),
\]
which gives question~\eqref{eqn:bayes}.

\section{LEMON}
\label{appendix:b}

Chinese Spelling Correction (CSC) in recent years makes a great stride, with many methods emerging and making impressive performances on general benchmarks like SIGHAN-2015. However, an ultimate CSC system must be able to cope with diverse domains and contexts simultaneously and offer appropriate error correction recommendations. We find that the current well-trained models on a single-domain still suffer from poor performances on multi-domain scenarios. The community is now in great need of another general benchmark to evaluate and study the generalization ability of a CSC system. We thus present \textit{LEMON}, \textit{a \textbf{l}arge-scal\textbf{e} \textbf{m}ulti-d\textbf{o}main dataset with \textbf{n}atural spelling errors}.

LEMON spans 7 domains, including game (GAM), encyclopedia (ENC), contract (COT), medical care (MEC), car (CAR), novel (NOV), and news (NEW). As opposed to prior work, where the typos are deliberately created on correct sentences, LEMON consists of 23 thousand examples with natural spelling errors picked from daily writing of human, which admittedly requires more annotation resources. Our idea is to stick close to the real human language distribution.

\begin{table}[]
\centering
\small
\begin{tabular}{l|c|c|c|c|c}
\toprule
             & \textbf{NE} & \textbf{NPE} & \textbf{SL} & \textbf{NEC} & \textbf{NEP} \\ \midrule
Game         & 400         & 155          & 33.0        & 1.16         & 133          \\
Encyclopedia & 3434        & 1712         & 39.8        & 1.28         & 1217         \\
Contract     & 1026        & 474          & 40.1        & 1.19         & 331          \\
Medical care & 2090        & 1053         & 39.3        & 1.33         & 674          \\
Car          & 3451        & 1762         & 43.6        & 1.35         & 1236         \\
Novel        & 6000        & 3014         & 36.3        & 1.13         & 5819         \\
News         & 5892        & 2946         & 25.1        & 1.11         & 1963         \\ \hline\hline
SIGHAN-15    & 1100        & 541          & 30.6        & 1.30         & 370          \\ \bottomrule
\end{tabular}
\caption{Data statistics for LEMON (NE: number of examples, NPE: number of positive examples, SL: sentence length, NEC: number of error characters per example,  NEP: number of edit pairs). SIGHAN-15 refers to the SIGHAN-15 test set.}
\label{stat}
\end{table}

LEMON contains a diverse collection of edit pairs and context, e.g. some cases requiring the domain-specific knowledge, some requiring the inference. This section presents a more concrete look at the examples in LEMON. For each case, we are going to demonstrate the source sentence, target sentence (human annotated), as well as the model prediction. As it turns out, the current model can hardly address those challenging cases.

\begin{CJK}{UTF8}{gkai}
\indent \textbf{Case 1: expertise (from MEC)}

\noindent $\bullet$ 头孢过敏可以用大环类酯。 「SRC」

\noindent $\bullet$ 头孢过敏可以用大环内酯。 「TRG」

\noindent $\bullet$ 头孢过敏可以用大环类酯。 「BERT」

A professional word 大环类酯 (\textit{macrolides antibiotics}) is misspelled here, which can be very hard to correct if the model is not exposed to specific knowledge during the training process.

\indent \textbf{Case 2: referential inference (from MEC)}

\noindent $\bullet$ 色盲眼镜是用于矫正色觉障碍的一种眼睛。 「SRC」

\noindent $\bullet$ 色盲眼镜是用于矫正色觉障碍的一种眼镜。 「TRG」

\noindent $\bullet$ 色盲眼镜是用于矫正色觉障碍的一种眼睛。 「BERT」

眼镜 (\textit{glasses}) is misspelled to 眼睛 (eyes) here. We notice that \textit{glasses} is mentioned earlier in the sentence, which requires the model to make the association based on the global context, albeit this is easy for human.

\indent \textbf{Case 3: unusual expression but globally correct (from GAM)}

\noindent $\bullet$ 但好像从没见人淘出过银两。 「SRC」

\noindent $\bullet$ 但好像从没见人淘出过银两。 「TRG」

\noindent $\bullet$ 但好像从没见人掏出过银两。 「BERT」

淘出 (\textit{find out}) is rarely expressed compared to 掏出 (\textit{take out}). The model is inclined to miscorrect those unusual expressions. Both \textit{find out coins} and \textit{take out coins} are correct expressions. According to the global context, however, we can know the background here is someone who digs for treasure. Hence, it should be \textit{found out} here.

\indent \textbf{Case 4: fixed pair (from ENC)}

\noindent $\bullet$ 可爱的动物共同构成了一副让人惊艳不已的画面。 「SRC」

\noindent $\bullet$ 可爱的动物共同构成了一幅让人惊艳不已的画面。 「TRG」

\noindent $\bullet$ 可爱的动物共同构成了一副让人惊艳不已的画面。 「BERT」

Since one will use 一副 \textit{a pair of} with 画面 (\textit{scene}), it should be corrected to 一幅 (\textit{a picture of}) here. However, there is a long attributive that separates them apart. The model fails to make it as a result.

\indent \textbf{Case 5: locally correct but globally incorrect expression (from CAR)}

\noindent $\bullet$ 发动机发生故障切记盲目拆检。 「SRC」

\noindent $\bullet$ 发动机发生故障切忌盲目拆检。 「TRG」

\noindent $\bullet$ 发动机发生故障切记盲目拆检。 「BERT」

切记 (\textit{remember}) and 切忌 (\textit{remember not}) are antonyms and both of them are correct expressions. According to the global context, what it means here is not to do something. Hence, \textit{remember} should be corrected to \textit{remember not}.

We can find that most of the cases here are expertise-free, but rather require more or less contextual comprehension and inference. Unfortunately, the current model is still weak in inference, perhaps more contextualized CSC methods could be developed in future study.

\indent \textbf{Case 6: multiple typos (from COT)}

\noindent $\bullet$ 由于上述原因试乙方无法履行保证时以方不承担责任。 「SRC」

\noindent $\bullet$ 由于上述原因使乙方无法履行保证时乙方不承担责任。 「TRG」

\noindent $\bullet$ 由于上述原因使乙方无法履行保证时以方不承担责任。 「BERT」

This case contains more than one errors.

\end{CJK}

\end{document}